\PassOptionsToPackage{usenames,dvipsnames,table}{xcolor}
\documentclass[11pt,a4paper]{article}
\usepackage[]{emnlp2018} 
\usepackage{times}
\usepackage{latexsym}
\usepackage{url}
\usepackage{helvet}
\usepackage{courier}
\usepackage{latexsym}
\usepackage{amssymb}
\usepackage{graphicx}
\usepackage[utf8]{inputenc} 
\usepackage[T1]{fontenc}    
\usepackage{url}
\usepackage{balance}  
\usepackage{mdwlist}
\usepackage{graphics}
\usepackage{color}
\usepackage{rotating}
\usepackage{booktabs}
\usepackage{epsfig}
\usepackage{alltt}
\usepackage{moreverb}
\usepackage{fancyvrb}
\usepackage{enumerate}
\usepackage{colortbl}           
\usepackage{xspace}
\usepackage{relsize}
\usepackage{array}
\usepackage{amsmath}
\usepackage{amsfonts}
\usepackage{txfonts}
\usepackage[labelfont=bf]{caption}
\usepackage{tabularx}
\usepackage{caption}
\usepackage{rotating}
\usepackage{latexsym}
\usepackage{multirow}
\usepackage{booktabs}
\usepackage{float}
\usepackage{array,etoolbox}
\usepackage{enumitem}
\usepackage{subfig}
\usepackage{colortbl}
\usepackage{arydshln}
\usepackage{setspace}
\usepackage{xcolor}
\usepackage[toc,page]{appendix}
\usepackage{soul}
\usepackage[draft]{pgf}
\usepackage{tipa}
\usepackage{bbm}

\usepackage[toc,page]{appendix}
\usepackage[labelfont=bf]{caption}

\definecolor{vlgray}{gray}{0.95}

\aclfinalcopy 
\def\aclpaperid{1735} 

\usepackage{array}
\newcolumntype{L}[1]{>{\raggedright\let\newline\\\arraybackslash\hspace{0pt}}m{#1}}
\newcolumntype{C}[1]{>{\centering\let\newline\\\arraybackslash\hspace{0pt}}m{#1}}
\newcolumntype{R}[1]{>{\raggedleft\let\newline\\\arraybackslash\hspace{0pt}}m{#1}}

\usepackage{lscape}
\newcolumntype{b}{>{\hsize=2.3\hsize}X}
\newcolumntype{s}{>{\hsize=.45\hsize}X}
\newcolumntype{m}{>{\hsize=.9\hsize}X}

\newcommand\T{\rule{0pt}{2.2ex}}       

\usepackage{etoolbox}
\newtoggle{draft}
\toggletrue{draft}
\iftoggle{draft}{
\newcommand{\tushark}[1]{\textcolor{BurntOrange}{[#1 \textsc{--tushar}]}}
\newcommand{\peterc}[1]{\textcolor{OliveGreen}{[#1 \textsc{--peter}]}}
\newcommand{\ashish}[1]{\textcolor{Maroon}{[#1 \textsc{--ashish}]}}
\newcommand{\dk}[1]{\textcolor{cyan}{[#1 \textsc{--dk}]}}
\newcommand{\fillthis}[1]{\textcolor{Red}{[#1]}}
}{
\newcommand{\tushark}[1]{}
\newcommand{\peterc}[1]{}
\newcommand{\ashish}[1]{}
\newcommand{\dk}[1]{}
\newcommand{\ed}[1]{}
\newcommand{\fillthis}[1]{}
}

\newcommand{\ensemble}{\textsc{Ensemble}\xspace}
\newcommand{\modular}{\textsc{NS}net\xspace}
\newcommand{\matcher}{Symbolic Matcher\xspace}
\newcommand{\lookup}{Symbolic Lookup\xspace}
\newcommand{\subf}{\ensuremath{h_i}\xspace}
\newcommand{\kbf}{\ensuremath{\mathit{kb}_j}\xspace}

\newcommand{\embover}{\textit{EmbOver}\xspace}
\newcommand{\embavg}{\textit{EmbAvg}\xspace}

\newcommand{\wordover}{\textit{WordOver}\xspace}

\newcommand{\ignore}[1]{}   

\usepackage{tabulary}

\newcommand{\xMapsto}[2][]{\ext@arrow 0599{\Mapstofill@}{#1}{#2}}
\def\Mapstofill@{\arrowfill@{\Mapstochar\Relbar}\Relbar\Rightarrow}


\newcolumntype{M}[1]{>{\centering\arraybackslash}m{#1}}

\usepackage{etoolbox}
\preto\align{\par\nobreak\normalsize\noindent}
\expandafter\preto\csname align*\endcsname{\par\nobreak\normalsize\noindent}

\title{Bridging Knowledge Gaps in Neural Entailment via Symbolic Models}

\author{
Dongyeop Kang$^1$\quad Tushar Khot$^2$\quad Ashish Sabharwal$^2$\quad Peter Clark$^2$ \\
$^1$School of Computer Science, Carnegie Mellon University, Pittsburgh, PA, USA \\
$^2$Allen Institute for Artificial Intelligence, Seattle, WA, USA\\
{\tt $\{$dongyeok$\}$@cs.cmu.edu}\quad {\tt $\{$tushark,ashishs,peterc$\}$@allenai.org}
}

\begin{document}
\maketitle
\begin{abstract}
Most textual entailment models focus on lexical gaps between the premise text and the hypothesis, but rarely on knowledge gaps. We focus on filling these knowledge gaps in the Science Entailment task, by leveraging an external structured knowledge base (KB) of science facts. 
Our new architecture combines standard neural entailment models with a knowledge lookup module. To facilitate this lookup, we propose a fact-level decomposition of the hypothesis, and verifying the resulting sub-facts against both the textual premise and the structured KB. 
Our model, \modular, learns to aggregate predictions from these heterogeneous data formats.
On the SciTail dataset, \modular outperforms
a simpler combination of the two predictions by 3\% and the base entailment model by 5\%.
\end{abstract}

\section{Introduction}
Textual entailment, a key challenge in natural language understanding, is a sub-problem in many end tasks such as question answering and information extraction. In one of the earliest works on entailment, the PASCAL Recognizing Textual Entailment Challenge, \citet{rte} define entailment as follows: text (or premise) P entails a hypothesis H if \emph{typically} a human reading P would infer that H is \emph{most likely} true. They note that this informal definition is ``based on (and assumes) common human understanding of language as well as \textit{common background knowledge}''.

While current entailment systems have achieved impressive performance by focusing on the language understanding aspect, these systems, especially recent neural models~\cite[e.g.][]{parikh2016decomposable,khot2018scitail}, do not directly address the need for filling knowledge gaps by leveraging common background knowledge.

\begin{figure}[t] 
\small
\centering
\fbox{
  \begin{minipage}[t]{\linewidth}
  \addtolength{\parskip}{1ex}

  \textbf{P}: The \colorbox{red!30}{aorta} is a \colorbox{blue!10}{large blood vessel} that \underline{moves} \underline{blood away from the heart} to the rest of the body.

  \textbf{H (entailed)}: \colorbox{red!30}{Aorta} is the \colorbox{orange!30}{major artery} \underline{carrying} recently oxygenated \underline{blood away from the heart}.

  \textbf{H' (not entailed)}: \colorbox{red!30}{Aorta} is the \colorbox{orange!30}{major vein} \underline{carrying} recently oxygenated \underline{blood away from the heart}.

\end{minipage}
}\vspace{0mm}
\caption{\label{fig:simple-example} Knowledge gap: Aorta is a major artery (not a vein). \emph{Large blood vessel} soft-aligns with \emph{major artery} but also with \emph{major vein}.}\vspace{-2mm}
\end{figure}

Figure~\ref{fig:simple-example} illustrates an example of P and H from SciTail, a recent science entailment dataset~\cite{khot2018scitail}, that highlights the challenge of \emph{knowledge gaps}---sub-facts of H that aren't stated in P but are universally true. 
In this example, an entailment system that is strong at filling lexical gaps may align \emph{large blood vessel} with \emph{major artery} to help conclude that P entails H. Such a system, however, would equally well---but incorrectly---conclude that P entails a hypothetical variant H' of H where \emph{artery} is replaced with \emph{vein}. A typical human, on the other hand, could bring to bear a piece of background knowledge, that aorta is a major artery (not a vein), to break the tie.

Motivated by this observation, we propose a new entailment model that combines the strengths of the latest neural entailment models with a structured knowledge base (KB) lookup module to bridge such knowledge gaps.
To enable KB lookup, we use a fact-level decomposition of the hypothesis, and verify each resulting sub-fact against both the premise (using a standard entailment model) and against the KB (using a structured scorer). The predictions from these two modules are combined using a multi-layer ``aggregator'' network.
Our system, called \modular, achieves 77.9\% accuracy on SciTail, substantially improving over the baseline neural entailment model, and comparable to the structured entailment model proposed by~\citet{khot2018scitail}.

\begin{figure*}[t]
\vspace{-1mm}
\small
\centering
{
\includegraphics[trim=0cm 5cm 11.5cm 0cm,clip,width=.90\linewidth]{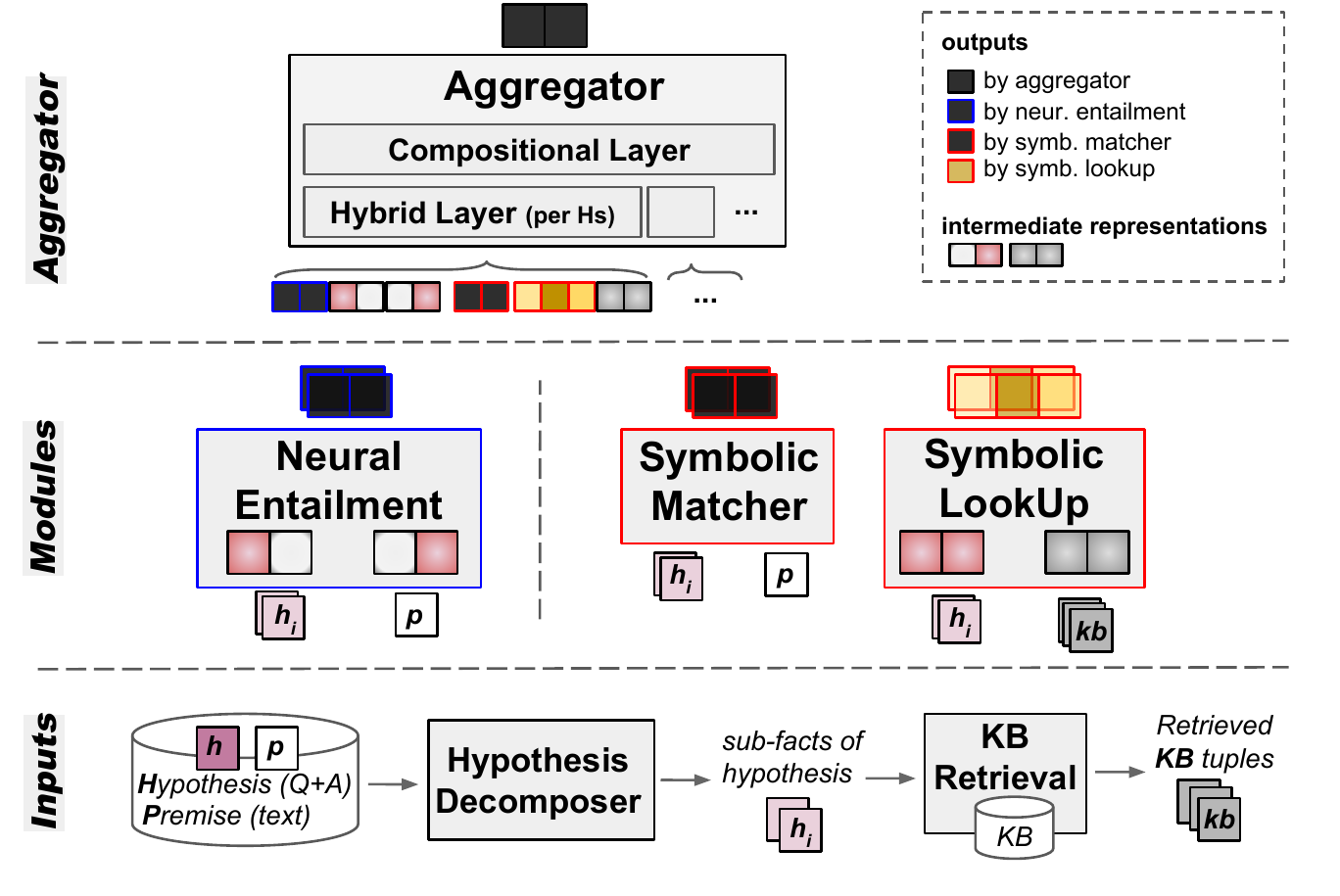}
}\vspace{-2mm}
\caption{\label{fig:nsl} Neural-symbolic learning in \modular. The bottom layer has QA and their supporting text
in SciTail, and the knowledge base (KB).
The middle layer has three modules: 
Neural Entailment (blue) and \matcher and \lookup (red).
The top layer takes the outputs (black and yellow) and intermediate representation from the middle modules, and hierarchically trains with the final labels.
All modules and aggregator are jointly trained in an end-to-end fashion. }
\vspace{-1mm}
\end{figure*}

\section{Neural-Symbolic Learning}

A general solution for combining neural and symbolic modules remains a challenging open problem. As a step towards this, we present a system in the context of neural entailment that demonstrates a successful integration of the KB lookup model and simple overlap measures, opening up a path to achieve a similar integration in other models and tasks. The overall system architecture of our neural-symbolic model for textual entailment is presented in Figure~\ref{fig:nsl}. 
We describe each layer of this architecture in more detail in the following sub-sections.

\subsection{Inputs}
We decompose the hypothesis and identify relevant KB facts in the bottom ``inputs'' layer (Fig.~\ref{fig:nsl}).

\paragraph{Hypothesis Decomposition:}
To identify knowledge gaps, we must first identify the facts stated in the hypothesis $h=(h_1,h_2..)$.
We use ClausIE~\cite{del2013clausie} to break $h$ into sub-facts. ClausIE tuples need not be verb-mediated and generate multiple tuples derived from conjunctions, leading to higher recall than alternatives such as Open IE~\cite{banko2007open}. \footnote{While prior work on question answering in the science domain has successfully used Open IE to extract facts from sentences~\cite{khot2017answering}, one of the key reasons for errors was the lossy nature of Open IE.}

\paragraph{Knowledge Base (KB):}
To verify these facts, we use the largest available clean knowledge base for the science domain~\cite{dalvi2017domain}, with 294K simple facts, as input to our system. The knowledge base contains subject-verb-object (SVO) tuples with short, one or two word arguments (e.g., hydrogen; is; element). Using these simple facts ensures that the KB is only used to fill the basic knowledge gaps and not directly prove the hypothesis irrespective of the premise.

\paragraph{KB Retrieval:}
The large number of tuples in the knowledge base makes it infeasible to evaluate each hypothesis sub-fact against the entire KB. Hence, we retrieve the top-100 relevant knowledge tuples, $K'$, for each sub-fact based on a simple Jaccard word overlap score. 

\subsection{Modules}
We use a Neural Entailment model to compute the entailment score based on the premise, as well as two symbolic models, \matcher and \lookup, to compute entailment scores based on the premise and the KB respectively (middle layer in Fig.~\ref{fig:nsl}). 
\paragraph{Neural Entailment}
We use a simple neural entailment model, Decomposable Attention~\cite{parikh2016decomposable}, one of the state-of-the-art models on the SNLI entailment dataset~\cite{bowman2015large}. However, our architecture can just as easily use any other neural entailment model. We initialize the model parameters by training it on the Science Entailment dataset. Given the sub-facts from the hypothesis, we use this model to compute an entailment score $n(\subf, p)$ from the premise to each sub-fact $\subf$.

\paragraph{\matcher}
In our initial experiments, we noticed that the neural entailment models would often either get distracted by similar words in the distributional space (false positives) or completely miss an exact mention of $\subf$ in a long premise (false negatives). To mitigate these errors, 
we define a \matcher model that compares exact words in $\subf$ and $p$,
via a simple asymmetric bag-of-words overlap score: 
\begin{align*}
m(\subf, p) = \frac{|\subf \cap p |}{|p|}
\end{align*}
One could instead use more complex symbolic alignment methods such as integer linear programming~\cite{khashabi2016question,khot2017answering}.

\paragraph{\lookup}
This module verifies the presence of the hypothesis sub-fact \subf in the retrieved KB tuples $K'$, by comparing the sub-fact to each tuple and taking the maximum score. Each field in the KB tuple $\kbf$ is scored against the corresponding field in \subf (e.g., subject to subject) and averaged across the fields. To compare a field,
we use a simple word-overlap based Jaccard similarity score, $\textit{Sim}(a, b) = \frac{|a \cap b|}{|a \cup b|}$. 
The lookup match score for the entire sub-fact and kb-fact is:
\begin{align*}
\mathit{Sim}_f(\subf, \kbf) = \left( \sum_k \textit{Sim}(\subf[k], \kbf[k]) \right) / 3
\end{align*}
and the final lookup module score for $\subf$ is:
\begin{align*}
l(\subf) = & \max_{\kbf \in K'} \textit{Sim}_f(\subf, \kbf) 
\end{align*}
Note that the \lookup module assesses whether a sub-fact of H is universally true. Neural models, via embeddings, are quite strong at mediating between P and H. The goal of the KB lookup module is to complement this strength, by verifying universally true sub-facts of H that may not be stated in P (e.g. ``aorta is a major artery'' in our motivating example). 

\ignore{
we consider three similarity scores: (1) Word-overlap Jaccard score \wordover; (2) cosine distance between averaged embeddings \embavg; and (3) a new similarity score, \embover, that combines the symbolic Jaccard score with cosine distance. To compute the \embover score, for each pair of words $(w_i, w_j)$ in a field $k$ of sub-fact $\subf$ and KB tuple \kbf, we compute their cosine distance. We then calculate the proportion of pairs with scores above a threshold, $\widehat{J}$:~\footnote{Empirically set to 0.9.}
\begin{align*}
 \frac{\sum_{w_i \in \subf, w_j \in \kbf}  \mathbbm{1}(\mathit{cosine}(w_i, w_j) > \widehat{J}) } {|\subf| * |\kbf|}
\end{align*}
}

\subsection{Aggregator Network}

For each sub-fact $\subf$, we now have three scores: $n(\subf, p)$ from the neural model, $m(\subf, p)$ from the symbolic matcher, and $l(\subf)$ from the symbolic lookup model. The task of the Aggregator network is to combine these to produce a single entailment score. However, we found that using only the final predictions from the three modules was not effective.
Inspired by recent work on skip/highway connections~\cite{he2016deep,srivastava2015highway}, we supplement these scores with intermediate, higher-dimensional representations from two of the modules.

From the \lookup model, we use the representation of each sub-fact $\subf^{\mathit{enc}} = \mathit{Enc}(\subf)$ obtained by averaging word embeddings~\cite{pennington2014glove}
and individual similarity scores over the top-100 KB tuples $\mathit{emb}_i = [\ldots,\textit{Sim}_f(\subf, \kbf), \ldots]$.
From the neural entailment model, we use the intermediate representation of both the sub-fact of hypothesis and premise text 
from the final layer (before the softmax computation), $n^{v}(\subf, p) = [v_1; v_2]$. 

We define a hybrid layer that takes as input a simple concatenation of these representation vectors from the different modules:
\begin{align*}
\mathit{in}(h_i, p) = & [ \subf^{\mathit{enc}}; l(\subf); m(\subf, p); n(\subf, p); \\
 & \mathit{emb}_i; n^{v}(\subf, p)]
\end{align*}
The hybrid layer is a single layer MLP for each sub-fact $h_i$ that outputs a sub-representation $\textit{out}_i=\textit{MLP}(\textit{in}(h_i, p))$. A compositional layer then uses a two-layer MLP over a concatenation of the hybrid layer outputs from different sub-facts, $\{h_1,\ldots,h_I\}$, to produce the final label, 
\begin{align*}
\mathit{label} = \mathrm{MLP}([\mathit{out}_1; \mathit{out}_2; \cdots \mathit{out}_{I}])
\end{align*}
Finally, we use the cross-entropy loss to train the Aggregator network jointly with representations in the neural entailment and symbolic lookup models, in an end-to-end fashion. We refer to this entire architecture as the \modular network.

To assess the effectiveness of the aggregator network, we also use a simpler baseline model, \ensemble, that works as follows. For each sub-fact \subf, it combines the predictions from each model using a probabilistic-OR function, assuming the model score $P_m$ as a probability of entailment. This function computes  the probability of at least one model predicting that \subf is entailed, i.e.
$P(\subf) = 1 - \Pi_m (1 - P_m)$ where $m \in n(h_i, p), m(h_i, p), l(h_i)$.
We average the probabilities from all the facts to get the final entailment probability.\footnote{While more intuitive, performing an AND aggregation resulted in worse performance (cf.\ Appendix~\ref{app:ensemble} for details).}

\begin{table}[t]
\setlength{\doublerulesep}{\arrayrulewidth}
\caption{\label{tab:scitail} Entailment accuracies on the SciTail dataset. 
\modular substantially improves upon its base model and marginally outperforms DGEM.}
\centering\vspace{0mm}
\begin{tabular}{@{}l|cc@{}}
\toprule
\textbf{Entailment Model} & \textbf{Valid.} & \textbf{Test}  \\ 
\midrule
\T Majority	classifier &	63.3 & 60.3		\\
DecompAttn (Base model) & 73.1 &	74.3 \\
DecompAttn + HypDecomp & 71.8	&72.7 \\
DGEM & \textbf{79.6} & 77.3 \\
\hline
\T \ensemble (this work) & 75.2 &	74.8 \\
\modular (this work) & 77.4 & \textbf{77.9}\\ 
\bottomrule
\end{tabular}\vspace{-1mm}
\end{table}

\section{Experiments}

We use the SciTail \textbf{dataset}\footnote{\url{http://data.allenai.org/scitail}}~\cite{khot2018scitail} for our experiments, which contains 27K entailment examples with a 87.3\%/4.8\%/7.8\% train/dev/test split. The premise and hypothesis in each example are natural sentences authored independently as well as independent of the entailment task, which makes the dataset particularly challenging. We focused mainly on the SciTail dataset, since other crowd-sourced datasets, large enough for training, contained limited linguistic variation~\cite{gururangan:artifacts} leading to limited gains achievable via external knowledge.

For \textbf{background knowledge}, we use version v4 of the aforementioned Aristo TupleKB\footnote{\url{http://data.allenai.org/tuple-kb}}~\cite{dalvi2017domain}, containing 283K basic science facts. We compare our proposed models to Decomposed Graph Entailment Model (DGEM)~\cite{khot2018scitail} and Decomposable Attention  Model (DecompAttn)~\cite{parikh2016decomposable}.


\subsection{Results}

Table~\ref{tab:scitail} summarizes the validation and test accuracies of various models on the SciTail dataset. 
 The DecompAttn model achieves 74.3\% on the test set but drops by 1.6\% when the hypotheses are decomposed. The \ensemble approach uses the same hypothesis decomposition and is able to recover 2.1\% points by using the KB. The end-to-end \modular network is able to further improve the score by 3.1\% and is statistically significantly (at p-value 0.05) better than the baseline neural entailment model.  
The model is marginally better than DGEM, a graph-based entailment model proposed by the authors of the SciTail dataset
We show significant gains over our base entailment model by using an external knowledge base, which are comparable to the gains achieved by DGEM through the use of hypothesis structure. These are orthogonal approaches and one could replace the base DecompAttn model with DGEM or more recent models~\cite{tay2017compare,yin2018end}.

\ignore{
\begin{table}[ht]
\caption{\label{tab:option} Scoring measures used for Lookup have similar performance with \wordover being best. }
\centering\vspace{-3mm}
\small
\begin{tabular}{@{}l|cc}
\modular & Valid & Test  \\
\hline
\T $\,\,\llcorner$Lookup (WordOver) &77.39 &	\textbf{77.94}\\
$\,\,\llcorner$Lookup (EmbAvg) & 77.21	&76.6\\
$\,\,\llcorner$Lookup (EmbOver) & \textbf{77.52}	&76.46
\end{tabular}\vspace{-1mm}
\end{table}
Table~\ref{tab:option} compares the three similarity measures for the KB lookup module. Surprisingly, simple \wordover achieves the highest score, indicating the benefit of symbolic matching against the KB.
}


\begin{table}[ht]
\caption{\label{tab:kb} Ablation: Both \lookup and \matcher have significant impact on \modular performance.}\vspace{-1mm}
\centering
\begin{tabular}{@{}l|cc@{}}
\toprule
 & \textbf{Valid.} & \textbf{Test}  \\
\midrule
\modular & 77.39  &	77.94\\
- \matcher & 76.46&	74.73 (\textbf{-3.21\%})\\
- \lookup & 75.95 & 75.80 (\textbf{-2.14\%}) \\ 
- Both & 75.10 & 73.98 (\textbf{-3.96\%})\\ 
\bottomrule
\end{tabular}\vspace{0mm}
\end{table}

In Table~\ref{tab:kb}, we evaluate the impact of the \matcher and \lookup module on the best reported model. As we see, removing the symbolic matcher, despite its simplicity, results in a 3.2\% drop. Also, the KB lookup model is able to fill some knowledge gaps, contributing 2.1\% to the final score. Together, these symbolic matching models contribute 4\% to the overall score.

\subsection{Qualitative Analysis}

\begin{table*}[htb]
\centering
\footnotesize
\renewcommand{\arraystretch}{1.5}
\caption{\label{tab:examples}Few randomly selected examples in the test set between symbolic only, neural only, \ensemble and \modular inference. The symbolic only model shows its the most similar knowledge from knowledge base inside parenthesis. The first two example shows when knowledge helps fill the gap where neural model can't. The third example shows when \modular predicts correctly while \ensemble fails.}\vspace{-1mm}

\begin{tabularx}{\linewidth}{|X|}
\hline
\textbf{Premise}:
plant cells possess a cell wall , animals never .\\
\textbf{Hypothesis}:
a cell wall is found in a plant cell but not in an animal cell .\\
\hline
\end{tabularx}
\begin{tabularx}{\linewidth}{|>{\hsize=.6\hsize}X|c|c|c|c|}
\textbf{Sub-fact of hypothesis} &\textbf{neural only} & \textbf{symbolic only} & \textbf{\ensemble} & \textbf{\modular} \\\hline
a cell wall is found in a plant cell but not in an animal cell & F(0.47) & T(0.07) (cell is located in animal) & T(0.50) & -\\\hline
\textbf{Prediction} (true label: \textbf{T} (entail))&  \textbf{F} & \textbf{T} & \textbf{T} & \textbf{T}\\
\hline 
\end{tabularx}
\newline\vspace*{5mm}\newline
\begin{tabularx}{\linewidth}{|X|}
\hline
\textbf{Premise}:
the pupil is a hole in the iris that allows light into the eye .\\
\textbf{Hypothesis}:
the pupil of the eye allows light to enter .\\
\hline
\end{tabularx}
\begin{tabularx}{\linewidth}{|>{\hsize=.6\hsize}X|c|c|c|c|}
\textbf{Sub-fact of hypothesis} &\textbf{neural only} & \textbf{symbolic only} & \textbf{\ensemble} & \textbf{\modular} \\\hline
the pupil of the eye allows light to enter & F(0.43) & T(0.12), (light enter eye) & T(0.50) & -\\\hline
\textbf{Prediction} (true label: \textbf{T} (entail))& \textbf{F} & \textbf{T} & \textbf{T} & \textbf{T}\\
\hline 
\end{tabularx}
\newline\vspace*{5mm}\newline
\begin{tabularx}{\linewidth}{|X|}
\hline
\textbf{Premise}:
binary fission in various single-celled organisms ( left ) .\\
\textbf{Hypothesis}:
binary fission is a form of cell division in prokaryotic organisms that produces identical offspring .\\
\hline
\end{tabularx}
\begin{tabularx}{\linewidth}{|>{\hsize=.6\hsize}X|c|c|c|c|}
\textbf{Sub-facts of hypothesis} &\textbf{neural only} & \textbf{symbolic only} & \textbf{\ensemble} & \textbf{\modular} \\\hline
binary fission is a form of cell division in prokaryotic organisms & F(0.49) & T(0.07) (binary fission involve division) & T(0.52) & -\\\hline
binary fission is a form & F(0.63) & T(0.1) (phase undergo binary fission) & T(0.66) & -\\\hline
a form of cell division in prokaryotic organisms produces identical offspring & F(0.46) & T(0.05) (cell division occur in tissue) & T(0.48) & -\\\hline
\textbf{Prediction} (true label: \textbf{T} (entail))& \textbf{F} & \textbf{T} & \textbf{F} & \textbf{T}\\

\hline 
\end{tabularx}\vspace{0mm}
\end{table*}

Figure~\ref{tab:examples} shows few randomly selected examples in test set.
The first two examples show cases when the symbolic models help to change the neural alignment's prediction (F) to correct prediction (T) by our proposed \ensemble or \modular models.
The third question shows a case where the \modular architecture learns a better combination of the neural and symbolic methods to correctly identify the entailment relation while \ensemble fails to do so.

\section{Related Work}

Compared to neural only~\cite{bowman2015large,parikh2016decomposable} or symbolic only~\cite{khot2017answering,khashabi2016question} systems, our model takes advantage of both systems, often called neural-symbolic learning~\cite{garcez2015neural}. 
Various neural-symbolic models have been proposed for question answering~\cite{liang2016neural} and causal explanations ~\cite{kang2017detecting}. We focus on end-to-end training of these models specifically for textual entailment. 

Contemporaneous to this work, \citet{chen2018natural} have incorporated knowledge-bases within the attention and composition functions of a neural entailment model, 
while \citet{kang18acl} generate adversarial examples using symbolic knowledge (e.g., WordNet) to train a robust entailment model. We focused on integrating knowledge-bases via a separate symbolic model to fill the knowledge gaps. 



\section{Conclusion}
We proposed a new entailment model that attempts to bridge knowledge gaps in textual entailment by incorporating structured knowledge base lookup into standard neural entailment models. Our architecture, \modular, can be trained end-to-end, and achieves a 5\% improvement on SciTail over the baseline neural model used here. The methodology can be easily applied to more complex entailment models (e.g., DGEM) as the base neural entailment model. Accurately identifying the sub-facts from a hypothesis is a challenging task in itself, especially when dealing with negation. Improvements to the fact decomposition should further help improve the model. 

\section*{Acknowledgments}
We thank the anonymous reviewers  as well as Luke Zettlemoyer, Matt Gardner, Oyvind Tafjord, Julia Hockenmaier and Eduard Hovy for their comments and suggestions.

\bibliographystyle{acl_natbib}
\bibliography{nsnet}

\clearpage
\begin{appendix}

\section{Model Hyper-parameters}
\label{app:hyper}

Our hyper-parameters tuned on development set are: Adam optimizer with learning rate 0.05, maximum gradient norm 5.0, batch size is 32, embedding size 300, and hidden layer size of feed-forward network is 200 with dropout rate 0.1.
The maximum vocabulary size is set to 30,000, but our dataset has a smaller vocabulary. The models are trained with either the original hypotheses or the sub-facts generated by ClausIE. 

We test dropout ratio from $0.5$, $0.75$, $0.9$ to $1.0$, and encoder with glove averaging or LSTM.
The maximum length of question and knowledge sentence is 25, and maximum length of supporting sentence in SciTail dataset is 40\footnote{Science entailment dataset has long premise sentences}.
The hidden size of hybrid layer is $50$, and hidden size of compositional layer is $50$ and $2$. 
The maximum number of knowledge per question is $100$ and maximum number of sub-question per question is $5$.
The average number of sub-questions per question decomposed by \cite{del2013clausie} is around $3.5$.
The learning rate is $0.05$ and maximum gradient norm is $5.0$ with Adam optimizer, and batch size is $32$.
We train our neural methods and \modular network up to $25$ epochs and choose the best model with validation and obtain accuracy on test set with the best model.

Based on the grid search over the hyperparameters, 
our best \ensemble model uses \textit{EmbOver} matcher on glove embeddings without tuple structure and probabilistic OR for hybrid decisions and averaging with $0.5$ threshold for compositional decisions.

The best \modular model uses \textit{WordOver} matcher on glove encoding with tuple structure, no dropout ratio and sub-question training with neural models. 

\section{Additional Experiments}

\begin{figure}[htb]
\vspace{0mm}
\centering
{
\includegraphics[trim=1.5cm 1.0cm 1.5cm 1.1cm,clip,width=.52\linewidth]{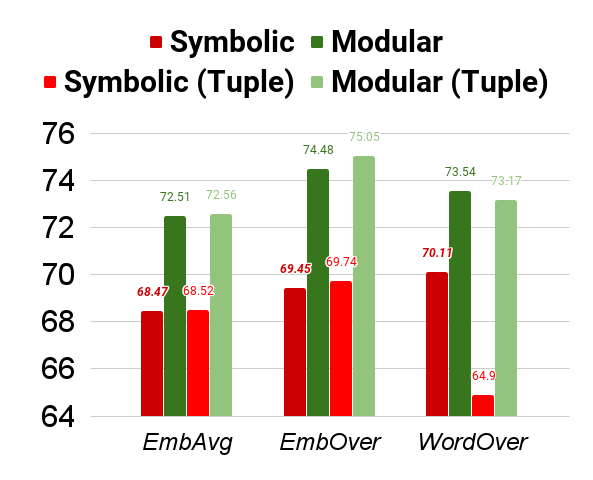}
\includegraphics[trim=0.4cm 0.6cm 0.4cm -0.8cm,clip,width=.42\linewidth]{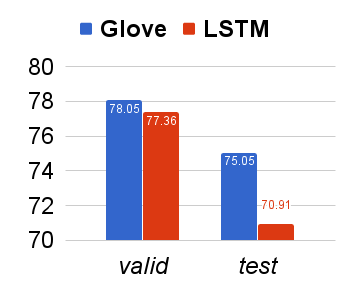}

}
\caption{\label{fig:matchers} Comparison between different matchers with tuplization (left) and different encoders (right).
}
\vspace{-2mm}
\end{figure}
For further analysis, we study effect of different matchers with(out) tuple structure, and different encoders (See Figure~\ref{fig:matchers}).
The left figure shows test accuracies of symbolic (red) and \modular (green) methods between three different lookup matchers (e.g., \textit{EmbeddingAverage}, \textit{EmbeddingOverlap}, \textit{WordOverlap}) and whether tuple structure is considered (light) or not (dark).
In most cases, \textit{EmbeddingOverlap} that takes advantages from \textit{EmbeddingAverage} and \textit{WordOverlap} outperforms the others, and tuple structure helps for finding best matching knowledge tuple in our world knowledge base.
The right figure shows accuracies between different encoders: averaging of glove word embeddings and LSTM with glove embedding initialization. 
LSTM is much worse in testing accuracy because of overfitting compared to glove embedding averaging.

\section{Observation on \ensemble Model Design}
\label{app:ensemble}
For the \ensemble network, we evaluated both OR and AND aggregation function and reported the best model. The use of AND is indeed intuitive. However, in addition to the empirical support for OR, the use of ClausIE to generate sub-facts makes probabilistic OR somewhat of a better fit, because of the following reason. ClausIE tries to generate every possible proposition in a sentence, erring on the side of higher recall at the cost of lower precision. This makes it unlikely for one to find good support for all generated sub-facts. This results in poor performance when using AND aggregation.

\end{appendix}

\end{document}